\documentclass[fleqn,11pt]{wlscirep}
\usepackage[utf8]{inputenc}
\usepackage[T1]{fontenc}
\usepackage{bm}
\usepackage{amsmath,amssymb}%
\usepackage{lineno}%
\usepackage{multirow}

\newcommand{\bi}{\begin{itemize}}
\newcommand{\ei}{\end{itemize}}
\newcommand{\ba}{\begin{array}}
\newcommand{\ea}{\end{array}}

\DeclareMathOperator*{\argmax}{arg\,max}

\newcommand{\avec}{\vec{a}}
\newcommand{\E}{\mathbb{E}}
\title{Data-driven simulator of multi-animal behavior with unknown dynamics via offline and online reinforcement learning} 

\author[1,2,*]{Keisuke Fujii}
\author[3,2]{Kazushi Tsutsui}
\author[4]{Yu Teshima}
\author[5]{Makoto Itoh}
\author[6]{Naoya Takeishi}
\author[7]{Nozomi Nishiumi}
\author[8]{Ryoya Tanaka}
\author[9]{Shunsuke Shigaki}
\author[10,2]{Yoshinobu Kawahara}
\affil[1]{Graduate School of Informatics, Nagoya University, Japan}
\affil[2]{RIKEN Center for Advanced Intelligence Project, Japan}
\affil[3]{Graduate School of Arts and Sciences, The University of Tokyo, Japan}
\affil[4]{Project team for SIP, Japan Agency for Marine-Earth Science and Technology, Japan}
\affil[5]{Faculty of Education, Shitennoji University, Japan}
\affil[6]{Graduate School of Engineering, The University of Tokyo, Japan}
\affil[7]{Graduate School of Science and Technology, Niigata University, Japan}
\affil[8]{Graduate School of Science, Nagoya University, Japan}
\affil[9]{Principles of Informatics Research Division, National Institute of Informatics, Japan}
\affil[10]{Graduate School of Information Science, The University of Osaka, Japan}

\affil[*]{Corresponding author: fujii@i.nagoya-u.ac.jp}

\begin{abstract}
Simulators of animal movements play a valuable role in studying behavior. Advances in imitation learning for robotics have expanded possibilities for reproducing human and animal movements. A key challenge for realistic multi-animal simulation in biology is bridging the gap between unknown real-world transition models and their simulated counterparts. Because locomotion dynamics are seldom known, relying solely on mathematical models is insufficient; constructing a simulator that both reproduces real trajectories and supports reward-driven optimization remains an open problem. We introduce a data-driven simulator for multi-animal behavior based on deep reinforcement learning and counterfactual simulation. We address the ill-posed nature of the problem caused by high degrees of freedom in locomotion by estimating movement variables of an incomplete transition model as actions within an RL framework. We also employ a distance-based pseudo-reward to align and compare states between cyber and physical spaces. Validated on artificial agents, flies, newts, and silkmoth, our approach achieves higher reproducibility of species-specific behaviors and improved reward acquisition compared with standard imitation and RL methods. Moreover, it enables counterfactual behavior prediction in novel experimental settings and supports multi-individual modeling for flexible what-if trajectory generation, suggesting its potential to simulate and elucidate complex multi-animal behaviors.
\end{abstract}

\begin{document}

\flushbottom
\maketitle
 
\section*{Introduction}
Advancements in computational modeling and data analysis have developed our ability not only to explore and interpret animal behavior but also to reproduce them.
This capacity to recreate observed behaviour under controlled conditions has become a cornerstone of modern animal-behaviour research.
For example, simulators that imitate animal movements can become key tools, as used in studies of rats and mice for laboratory courses \cite{corte2021anatomical}, modeling the behavior of nematodes \cite{mori2022probabilistic}, ants, and birds \cite{wijeyakulasuriya2020machine}, and also applied to issues that arise between animals and human society, such as bird strikes \cite{quaglietta2019simulating}. The development and use of this simulator are of considerable importance to both theoretical research and practical applications in various fields, including ethology, neuroscience, and robotics \cite{mori2022probabilistic}.
Moreover, by elucidating the causal mechanisms that govern collective dynamics, such simulators provide biological insights that are difficult to obtain through observation alone.

These simulators offer multiple advantages that enhance both theoretical and practical aspects of animal behavior research. By clearly representing animal behavior in controlled conditions, the simulator strengthens our fundamental understanding of behavioral patterns. Additionally, it facilitates a valuable exchange between scientific research and real-world applications, enabling the study and refinement of practical issues within laboratory settings before applying the findings to inform behavior-change practices. This ongoing interplay promotes the development of specialized solutions that effectively address real-world challenges \cite{corte2021anatomical, peng2020learning, wijeyakulasuriya2020machine, quaglietta2019simulating, mori2022probabilistic}. Furthermore, such a simulator is particularly useful in comprehending the actions of unrestrained animals that exhibit spontaneous and self-initiated movements, thereby contributing to a more comprehensive understanding of ethologically relevant and ecologically valid behaviors \cite{niv2021primacy}.

Building on these foundational capabilities, a significant challenge persists in simulating realistic multi-animal behaviors within biological sciences: bridging the gap between unknown transition models in simulated environments and their real-world counterparts. Traditional simulation approaches often rely on predefined rules or simplistic models that fail to capture the intricate dynamics of natural animal behaviors. 
However, many terrestrial animals exhibit abrupt stops and sudden starts of motion that cannot be captured by concise mathematical formulations, making such behaviors particularly difficult to capture with rule‑based or low‑order dynamic models.
This discrepancy, known as the domain gap, restricts the ability to replicate and analyze behaviors observed in real-world settings. Specifically, when dealing with biological multi-agents, the complexities of real-world dynamics are challenging to represent within the rigid frameworks of conventional simulators, resulting in limited applicability and reduced fidelity in behavioral representations.

Motivated by the need to address this domain gap, our research introduces a new data-driven simulator of multi-animal behavior utilizing deep reinforcement learning (RL). Recent advancements in RL, particularly those leveraging neural network architectures, have demonstrated remarkable flexibility and diversity in modeling complex behaviors within cyberspace \cite{ross2010efficient, ho2016generative, hester2018deep}. However, these advancements primarily cater to scenarios where the source dynamics are well-defined, such as in Sim-to-Real transfer learning \cite{rusu2017sim}, where knowledge is transferred from simulated environments to real-world applications like robotics \cite{schaal1996learning, kolter2007hierarchical}. In contrast, our work addresses a Real-to-Sim domain adaptation problem, where the source environment comprises real-world data with unknown and often intricate dynamics, and the target environment is a simulated cyberspace \cite{fujii2024adaptive}. This approach necessitates overcoming the absence of explicit transition models, which are typically unavailable in real-world scenarios, thereby presenting an ill-posed problem characterized by high degrees of freedom in animal locomotion.

\begin{figure}[!t]
\centerline{\includegraphics[width=1 \textwidth]{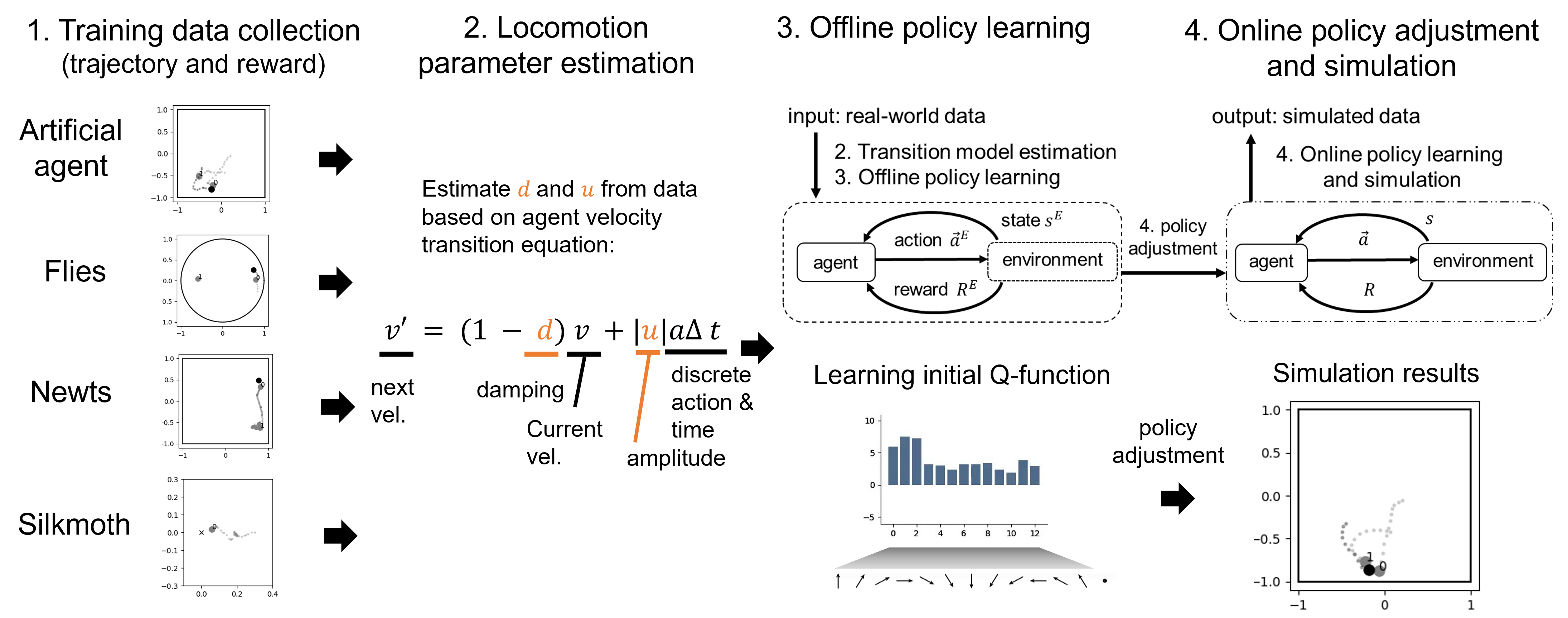}}
\caption{{\bf Schematic diagram of AnimaRL framework.}
The diagram illustrates the framework for modeling and simulating multi-agent behaviors, using both artificial and biological agents as demonstrations. Step 1 is training data collection, showing pursuit and escape trajectories of artificial agents, flies, newts, and silkmoth on a two-dimensional plane. The primary objective for pursuing agents is to make contact with the target, which acts as a reward signal in reinforcement learning. Step 2 is locomotion parameter estimation, where damping coefficient ($d$) and amplitude of discrete input ($u$) are estimated as fundamental components in the agent velocity transition function. Step 3 is offline policy learning, where the initial policy (in this case, Q-function) is learned from demonstration data without simulation. Step 4 is online policy adjustments. Each agent's policy is refined via interaction with environments (i.e., reinforcement learning with simulation). This integrated approach allows for a more realistic simulation of multi-agent behaviors.
}
\label{fig:diagram}
\end{figure}

To overcome these challenges, we propose AnimaRL (animal simulator with a deep RL) framework that estimates the locomotion parameters of the transition model and learns an offline policy and adjusts an online policy within RL paradigm as illustrated in Fig. \ref{fig:diagram}. AnimaRL allows the simulator to autonomously learn and adapt to the nuanced behaviors of multiple animals from extensive real-world datasets. Additionally, we incorporate a distance-based pseudo-reward mechanism called Deep Q-learning with Distance-based Imitation Learning (DQDIL).
Our framework recovers interpretable locomotion parameters and compute state-action values, allowing researchers to inspect both the learned dynamics and decision process.
It aligns and compares states between the cyber and physical spaces of the agents, ensuring consistency and enhancing the realism of simulated behaviors. Our approach has been rigorously validated using diverse datasets encompassing flies, silkmoths, and newts. The results demonstrate superior reproducibility of species-specific behaviors and reward acquisition compared to existing imitation learning techniques. Moreover, the simulator's capacity for counterfactual behavior prediction with a variant called DQCIL (Counterfactual Imitation Learning) in different conditions for agents and silkmoth, emphasizes its versatility and potential for facilitating virtual experiments in previously untested experimental settings. By enabling flexible movement trajectories and accommodating multiple individuals, our framework not only bridges the Real-to-Sim domain gap but also contributes a deeper understanding of complex multi-animal behaviors, thereby advancing both theoretical research and practical applications across ethology, neuroscience, and robotics.

\section*{Results}
\subsection*{AnimaRL framework.}

To address the challenges of simulating realistic multi-animal behaviors, we developed AnimaRL, a new framework that integrates deep RL with data-driven modeling (Fig. \ref{fig:diagram}). The architecture of AnimaRL comprises several key modules: locomotion parameter estimation, offline policy learning, online policy adjustment, and simulation environment interface. The framework begins with the input of real-world animal behavioral data (trajectories and rewards) to ensure compatibility with the RL algorithms. AnimaRL employs Deep Q-Network (DQN) and distance-based pseudo-reward (called DQDIL) to learn integrated reward and imitation policies from the processed data. These policies govern the simulated agents within the virtual environment, enabling them to exhibit both earning-reward multi-agent interactions and imitating the demonstration data in real-world scenarios.

After data collection in Step 1 of Fig. \ref{fig:diagram} (see Materials and Methods Section), Step 2 in the AnimaRL framework is \textit{locomotion parameter estimation}, which involves inferring the damping coefficient ($d$) and the amplitude of control input ($u>0$) from observed position and velocity data of animal agents. This estimation is governed by the \textit{agent velocity transition equation}:

\begin{equation}
v' = (1 - d) v + ua \Delta t,
\label{eq:velocity_transition}
\end{equation}
where $v$ represents the current velocity of an animal agent $([m\,s^{-1}])$, $d$ is the damping coefficient that accounts for resistance or frictional effects $([a.u.])$, $u$ denotes the amplitude of discrete control input $a$ (a unit vector) driving the agent's movement $([m\,s^{-2}])$, and $\Delta t$ is the discrete time increment $([s])$. Unlike traditional motion equations used in physics, this transition model operates within the RL paradigm, regarding $d$ and $u$ as parameters to be learned from demonstration data before the RL.
In addition, although the spatial unit is defined as meters for clarity, all spatial coordinates are actually normalized and therefore dimensionless in the experiments.


Following parameter estimation, AnimaRL employs a two-phase reinforcement learning approach to optimize and adapt behavior policies (for details, see Materials and Methods Section). Step 3, \textit{offline policy learning}, utilizes historical behavioral data to train the RL agents without real-time interaction. During this phase, we apply Q-learning algorithms to develop integrated reward and imitation policies that simultaneously maximize the original reward functions and mimic the observed data trajectories. This dual-objective optimization ensures that the agents not only achieve high performance in terms of reward acquisition but also replicate the movement patterns observed in real-world multi-animal interactions.

Subsequently, as Step 4, \textit{online policy adaptation} allows the agents to refine their integrated reward and imitation policies through online interactions within the simulated environment. This step leverages adaptive learning mechanisms to further enhance both reward optimization and behavior imitation, enabling the agents to dynamically respond to changing conditions and interactions. The combination of offline and online RL phases ensures that AnimaRL not only replicates existing behaviors with high fidelity but also adapts to novel scenarios, enhancing the simulator's versatility and robustness.

\subsection*{Examples of simulation results.}
\begin{figure}[!t]
\centerline{\includegraphics[width=1 \textwidth]{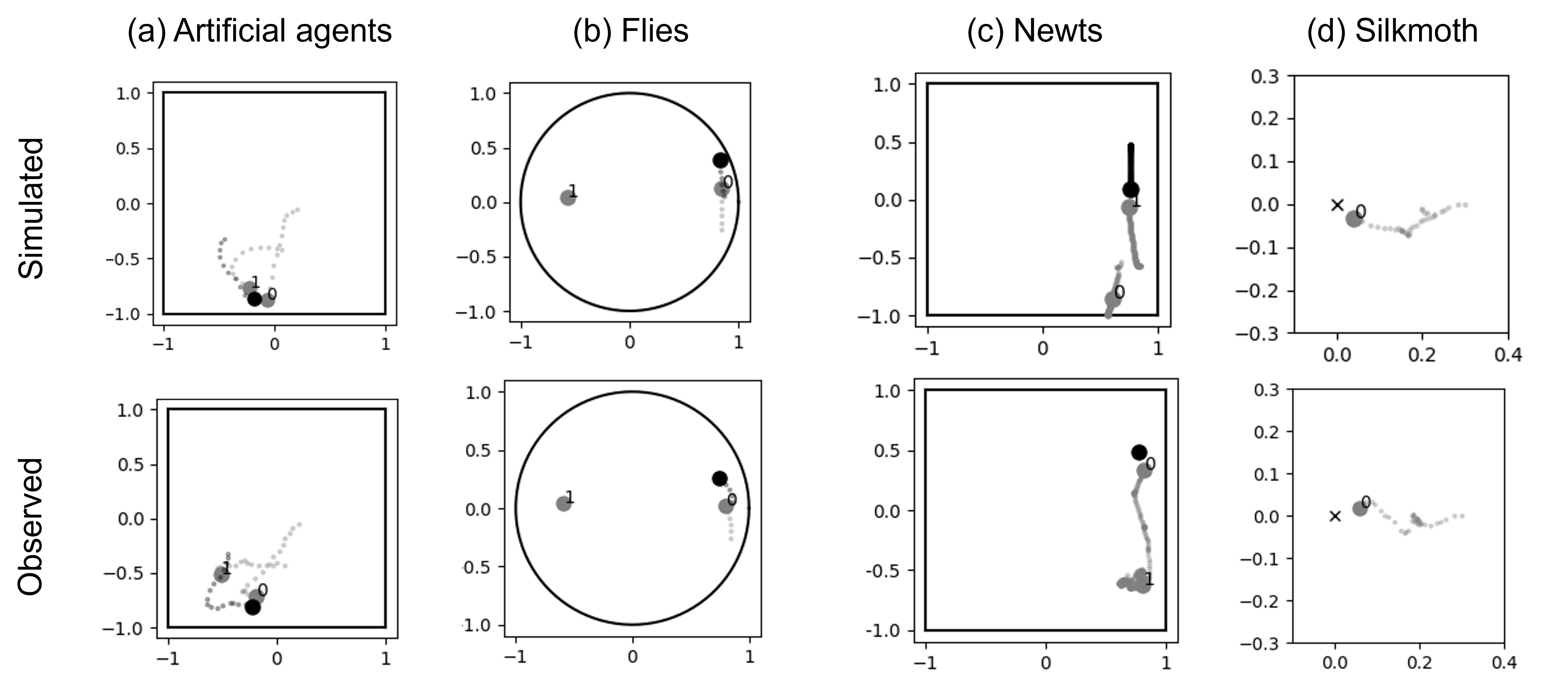}}
\caption{{\bf Simulated (top row) versus observed (bottom row) trajectories for agents and animals.}
Panels illustrate a single two-agent trial for (a) artificial agents, (b) flies, (c) newts, and (d) silkmoths. Solid outlines denote the physical arena used for data collection: square for agents and newts, circular for flies; the field for the silkmoth had no enclosing wall. In panels a-c a single black marker (the agent being chased or females) is pursued by two gray markers 0 and 1 (chasing agents or males). In panel d the silkmoth (gray) moves toward a stationary odor source indicated by an ``x''. Axes are normalized so that the arena center is at the origin. Histories begin at the start location and extend to a moment just before the target (black individual or odor source) is reached.
}
\label{fig:example}
\end{figure}

We first present examples of simulation results with pre-trained DQDIL (both offline and online RL) for the artificial agents, flies, newts, and silkmoth (Fig. \ref{fig:example}). 
All datasets include positional data in a two-dimensional plane and the detailed dataset descriptions are explained in the Materials and Methods section. 
The artificial agent dataset consisted of trajectories generated from a predator-prey simulation task in a simplified virtual environment. Specifically, two predator agents aimed to capture a prey agent within a square space, where the positions and velocities were continuously tracked at discrete time intervals. Each episode began with randomized initial positions, and rewards were given when predators successfully captured the prey or when the prey evaded capture for the duration of the simulation. This dataset served as a controlled baseline to validate the effectiveness of the proposed method against scenarios with known dynamics.
At the stage of locomotion parameter estimation in Fig. \ref{fig:diagram}, the damping parameters $d$ and the amplitudes of discrete input $u$ are estimated and example results are shown in Table \ref{tab:param_estimates} and Fig. \ref{fig:example}a, respectively.
The observed trajectories seem to be convergent as in the simulation, differing only by small lateral shifts near their final positions.
Quantitatively, $d$ and $u$ were accurately estimated and root mean squared errors (RMSEs) in the estimation and observation of next step velocity were less than 0.045 for all agents, suggesting reasonable fitting as the locomotion model. 

\begin{table}[t]
  \centering
  \caption{Locomotion parameter estimation and validation.}
  \label{tab:param_estimates}
  \small
  \begin{tabular}{llccccc}
    \toprule
    \multirow{2}{*}{Agent/Animal} & \multirow{2}{*}{Parameters/error} &
    \multicolumn{2}{c}{Artificial agents} & Flies & Newts & Silkmoth \\
    \cmidrule(lr){3-4} \cmidrule(lr){5-7}
    & & Estimated & Ground Truth & Estimated & Estimated & Estimated \\
    \midrule
    \multirow{3}{*}{0 (gray)}
      & Damping $d$           & 0.254 & 0.250 & 0.013 & 0.011 & 0.208 \\
      & Input amplitude $u$   & 2.957 & 3.000 & 0.143 & 0.013 & 0.021 \\
      & Velocity RMSE         & 0.043 & ---   & 0.076 & 0.022 & 0.009 \\
    \addlinespace
    \multirow{3}{*}{1 (gray)}
      & Damping $d$           & 0.258 & 0.250 & 0.017 & 0.010 & ---   \\
      & Input amplitude $u$   & 2.979 & 3.000 & 0.165 & 0.014 & ---   \\
      & Velocity RMSE         & 0.040 & ---   & 0.056 & 0.013 & ---   \\
    \addlinespace
    \multirow{3}{*}{2 (black)}
      & Damping $d$           & 0.257 & 0.250 & 0.014 & 0.014 & ---   \\
      & Input amplitude $u$   & 3.000 & 3.000 & 0.156 & 0.014 & ---   \\
      & Velocity RMSE         & 0.044 & ---   & 0.105 & 0.008 & ---   \\
    \bottomrule
  \end{tabular}
\end{table}

In the real-world animal datasets, the fly dataset comprised trajectories of two male flies actively pursuing a female fly within a circular space, reflecting male-female interactions characteristic of natural courtship behavior.
Newt dataset included trajectories of two male newts pursuing a female newt within a rectangular area but normalized to a square space.
Silkmoth dataset included trajectory data from a male silkmoth in a virtual reality environment simulating multisensory cues (odor, vision, wind), tracking navigation behavior toward an odor source \cite{yamada2021multisensory}.
For the stage of locomotion parameter estimation, the damping parameters $d$ and the amplitudes of discrete input $u$ are estimated and example results are shown in Table \ref{tab:param_estimates} and Figs. \ref{fig:example}b, c, and d.
The silkmoth and newt models performed better (0.009 and less than 0.023, respectively) than artificial agent models, whereas the fly model showed a larger error of up to 0.105. The following section examines how these accuracy differences influence the simulation outcomes. 

\begin{figure}[!t]
\centerline{\includegraphics[width=1 \textwidth]{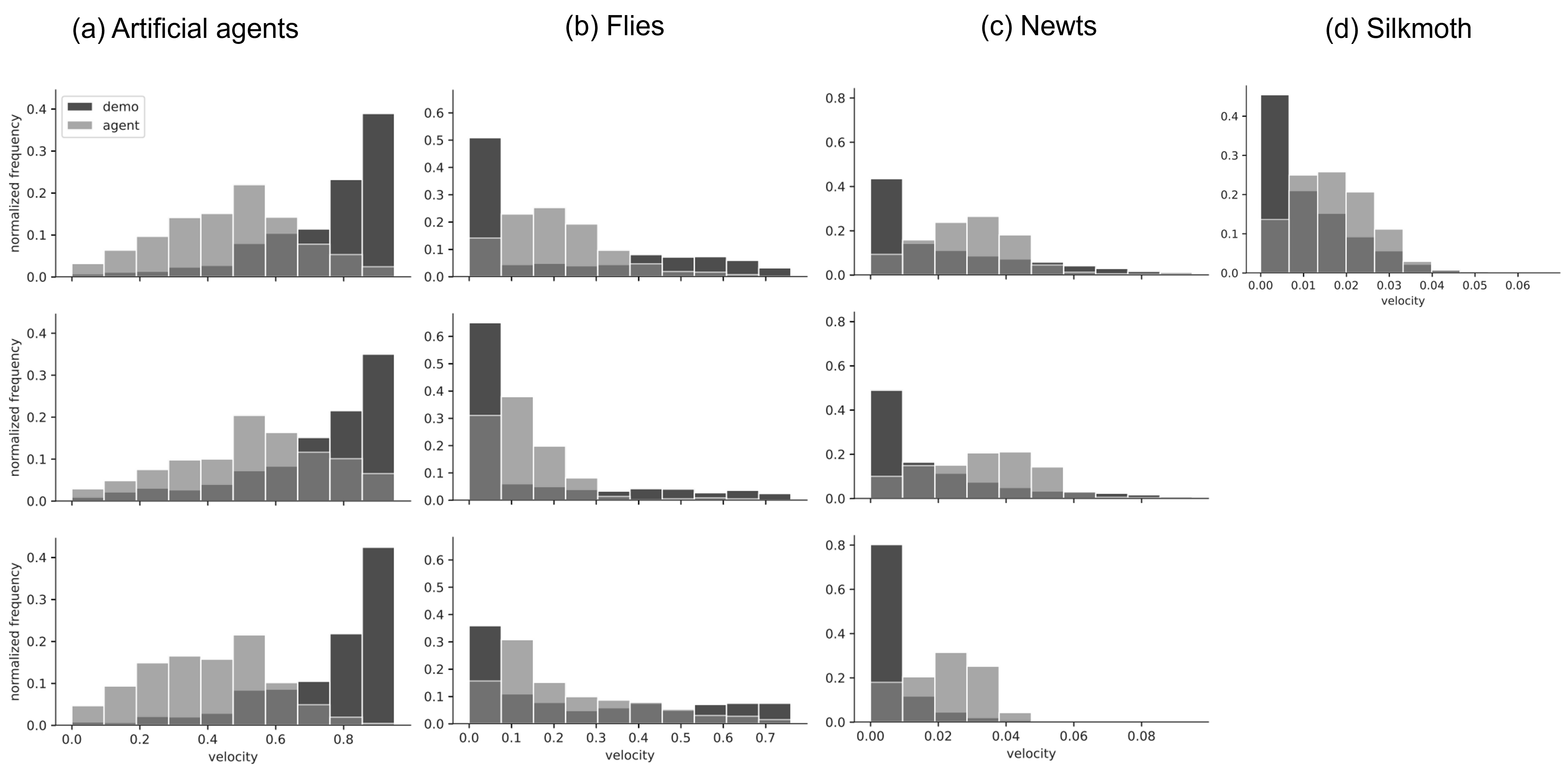}}
\caption{{\bf Histograms of instantaneous velocity magnitudes.}
For (a) artificial agents, (b) flies, (c) newts, and (d) silkmoths, stacked bar histograms compare the normalized frequency of velocities from demonstration data (dark bars) with those produced by the pre-trained DQDIL (light bars). Bin widths were chosen so that the areas sum to 1.0 within every panel. Artificial agents exhibit a high-velocity dominance, while flies, newts, and silkmoth show low-velocity dominance characteristic of the demonstrations. 
}
\label{fig:velocity}
\end{figure}

Figure \ref{fig:velocity} shows the measured and simulated velocity distributions across species. In the artificial agents, the observation data were biased toward higher velocities, and the fly, newt, and silkmoth datasets are dominated by near-zero velocity.
Both characteristics are challenging in learning-based simulators because sustaining high velocity demands continual acceleration in a single direction, whereas remaining at zero velocity earns no task reward (i.e., the agent makes no progress toward the target).
In particular, the fly histogram is the most problematic: it combines an excessive spike at rest with a long right tail, so any single-mode parametric model struggles to match both extremes simultaneously. 
This dual mismatch may be related to the larger velocity RMSE for flies and suggests that future work should incorporate mixture or state-switching dynamics when modeling biological locomotion.

\subsection*{Reinforcement learning model verification.}
\begin{figure}[!t]
\centerline{\includegraphics[width=1 \textwidth]{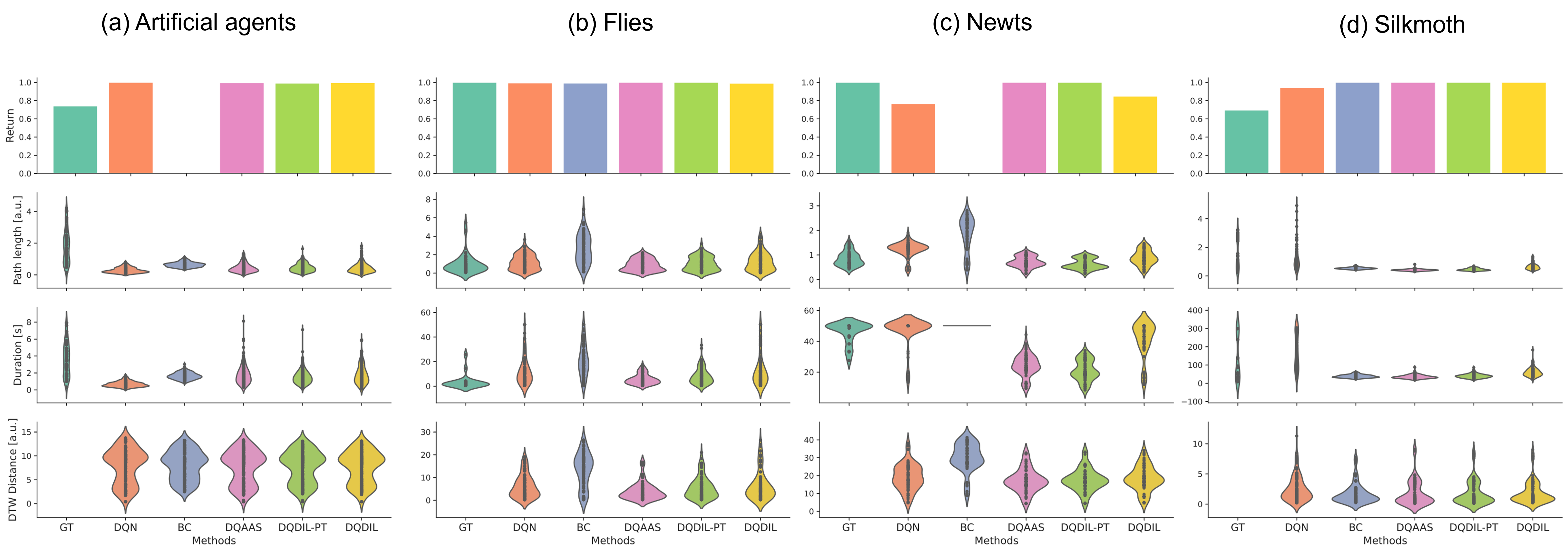}}
\caption{{\bf Quantitative comparison of generated trajectories across learning methods.}
Violin plots summarize (top row) episode duration and (bottom row) dynamic time warping (DTW) distance between generated and demonstration trajectories for (a) artificial agents, (b) flies, (c) newts, and (d) silkmoth. Columns correspond to ground truth (GT) and five learning baselines: deep Q-network (DQN), behavioral cloning (BC), deep Q adaptive action supervision (DQAAS), pre-trained distance-based imitation learning (DQDIL-PT), and DQDIL without pretraining. Wider violins indicate higher density of runs at that metric value; black dots denote individual trials. 
}
\label{fig:models}
\end{figure}

To verify the model's performance (both reproducibility for the imitation and generalization ability for obtaining rewards) across both the controlled (artificial agents) and real-world (flies, newts, silkmoth) cases, we evaluated obtained returns, path lengths, total duration, and distance from demonstrations (Fig. \ref{fig:models}). 
For a common metric of generalization ability in each species, we quantified ``return'' as the number of successful contacts with the target: either the count of episodes in which one of the two chasers or males reached the evader or female for the agent, fly, and newt datasets, or the count of odor-source arrivals for silkmoths. As reproducibility metrics, path lengths and total duration within an episode were examined (the former is defined as the mean path length of the two chasers, two males, or the single silkmoth). 
Since one of our goals was to judge how closely a learned policy reproduces ground truth (GT), we compared the kernel-density estimates (KDEs) of the path length and duration against the corresponding GT KDEs and used the resulting kernel-density distances as distribution similarity measures.
Finally, we assessed spatial fidelity with the dynamic time warping (DTW) distance between simulated and GT trajectories, testing improvements relative to behavioral cloning (BC) by paired bootstrap resampling. 
For the three metrics, we also performed the paired bootstrap contrasts of methods by computing the difference between the GT (or BC) and each method.
Note that since the bootstrap ANOVA consistently yielded positive F-statistics, we followed up with paired bootstrap contrasts, but to avoid the inflation of family-wise error inherent in exhaustive multiple testing, we limited these post-hoc analyses to a priori comparisons of primary scientific interest only.

We treated the DQN baseline only as a qualitative reference because it is trained without demonstrations and therefore not strictly comparable. Recognizing that higher return (here, reaching a target) can come at the cost of trajectory realism, we did not seek to crown a single best model. Instead, we used bootstrap 95\% confidence intervals on the GT-difference metrics to pinpoint where the previously proposed DQAAS (Deep-Q learning with adaptive action supervision) \cite{fujii2024adaptive}, the pre-trained DQDIL-PT, and the DQDIL without offline RL (i.e., without pre-training). We computed these non-parametric confidence intervals to traditional p-value tests because they report both effect magnitude and uncertainty without imposing parametric assumptions or forcing binary accept/reject decisions.
 
\textit{Artificial agents.}
Figure \ref{fig:models}a shows that RL policies allowed each chaser to touch the evader in almost every rollout, so the return histogram was effectively saturated at 1.0. Since the chasers accelerate almost immediately toward the target, all three methods (DQAAS, DQDIL-PT, and DQDIL) produced markedly shorter paths than the ground truth. KDE gaps between the GT and the models were uniformly positive and substantial (95\% CI lower bounds > 1.18, medians > 1.283).
For episode duration, the KDE gap confidence interval for DQDIL (median = 0.811, 95\% CI = [0.687, 0.935]) lies entirely below that for the pre-trained variant DQDIL-PT (1.064, [0.944, 1.183]), demonstrating that DQDIL reproduces ground-truth timing more faithfully than its pre-trained counterpart. By contrast, the interval for DQAAS (0.882, [0.757, 1.008]) overlaps the DQDIL range, so the available bootstrap evidence does not allow us to claim a reliable difference between these two methods.
Across all three learning strategies the DTW distances tended to fall below the BC baseline: the bootstrap median reductions were 0.463-0.523 units, and every 95\% confidence interval lay mostly on the positive side of zero (lower bounds as high as -0.089, upper bounds as low as 1.036, but similar in the three methods), indicating a general (though not uniformly decisive) downward shift relative to BC.

\textit{Flies.}
In the fly dataset, as shown in Fig. \ref{fig:models}b, even BC attains almost perfect returns, reflecting the fact that the female can be easily reached. 
The challenge instead lies in reproducing the short, hesitant advance of the males. 
As seen in the second panel of Fig. \ref{fig:models}b, all three methods obtained path length distributions that diverged from the ground truth, with KDE-gap medians in the 0.258-0.413 range and 95\% CI lower bounds no smaller than 0.162.  Although the DQAAS traces looked qualitatively closer to GT in some episodes, every pairwise comparison showed a confidence interval spanning zero. 
About durations, none of the learned policies matched the GT episode durations: every KDE-gap confidence interval lay wholly above zero. However, the ordering of those gaps was consistent: DQAAS was closest to GT (median = 0.528, 95\% CI [0.373, 0.758]), DQDIL-PT came next (0.714, [0.529, 0.943]), and DQDIL was farthest (0.837, [0.664, 1.058]). The pairwise bootstrap results with the difference from the GT show that DQAAS outperformed both DQDIL-PT (95\% CI [-3.258, -0.421]) and DQDIL ([-8.991, -4.197]), while DQDIL-PT was closer to GT than DQDIL ([-7.293, -2.271]). 
The DTW distances in the three methods were smaller than the BC baseline. Bootstrap comparisons show median reductions of 9.280, 7.970, and 6.842 units for DQAAS, DQDIL-PT, and DQDIL, respectively, with 95\% CIs that lay entirely above zero in every case (lower bounds $\geq$ 4.929). The paired comparisons of the difference from GT indicate that DQAAS was significantly closer to the GT than both DQDIL-PT (95\% CI [-2.544, -0.135]) and DQDIL ([-3.904, -0.990]), whereas the gap between the two DIL variants was inconclusive (CI [-2.669, 0.449]). 

\textit{Newts.}
The newt data emphasize intermittent pauses, making pure reward maximization less informative. Consequently, DQDIL accepts a smaller contact-rate return but delivers trajectories that resemble GT more faithfully in space and time. 
Every method’s KDE gap versus GT in path length and duration remained above zero (median discrepancies were 0.258-0.602 and 0.407-1.358, respectively). 
Pairwise bootstrap results in the path length difference from GT and DTW also produced CIs that straddled zero for every comparison, with absolute medians less than 0.017 and 1.146, respectively.  
However, pair-wise results in durations show that DQDIL was significantly closer to the GT distribution than DQAAS (95\% CI = [8.679, 14.473]) and DQDIL-PT ([11.752, 17.601]), while DQAAS itself outperformed DQDIL-PT ([-5.847,-0.439]).

\textit{Silkmoths.}
For silkmoths, all methods reached the odor source with similar frequency. 
Kernel-density gaps for path lengths and duration imply that DQDIL reproduced GT most closely, its moving-distance and duration deviations (95\% CI: [0.947, 1.229] and [0.790, 1.036]) were smaller than those of DQAAS ([1.464, 1.629] and [1.152, 1.342]) or DQDIL-PT ([1.372, 1.540] and [1.094, 1.291]).
However, every contrast had a confidence interval that straddled zero, indicating no reliable pairwise advantage.
The same lack of separation was observed for DTW gaps relative to BC and all contrast intervals again overlapped zero. Thus, while KDE magnitudes suggest a slight edge for DQDIL, the contrast tests show that, statistically, the three methods are indistinguishable on path length, duration, and DTW.

\subsection*{Counterfactual prediction.}
To probe whether our model can adapt to unobserved conditions, we augmented our DQDIL with a counterfactual prediction head that predicts a binary cue specifying the experimental condition and is trained with an adversarial (negative-gradient) loss to reduce the sampling bias of the conditions \cite{fujii2024estimating}.
The resulting model is denoted DQCIL (Deep-Q Counterfactual Imitation Learning). 
In the artificial-agent task, the two conditions correspond to chasers that do not share reward (Condition 1) and chasers that do share reward (Condition 2) \cite{tsutsui2023synergizing,tsutsui2022emergence}. 
Since the DTW distance can be less interpretable, we compared the path length of GT, DQDIL, DQCIL for each condition, and counterfactual predictions when queried with the opposite cue (1 $\rightarrow$ 2 or 2 $\rightarrow$ 1). 
We performed the same procedure for real silkmoths, defining the full sensory condition (odor, wind, and vision; Condition 1) and a partial condition in which wind and vision are blocked (Condition 2).  
Since we expect that pretraining to stabilize learning in the synthetic-agent domain, where data and reward signals are sparse, but to be unnecessary in the richly sensed silkmoth task, we here show pretrained results only for agents and omitted it for silkmoths (for counterpart results, see Supplementary materials Text S1). 

\begin{figure}[!t]
\centerline{\includegraphics[width=1 \textwidth]{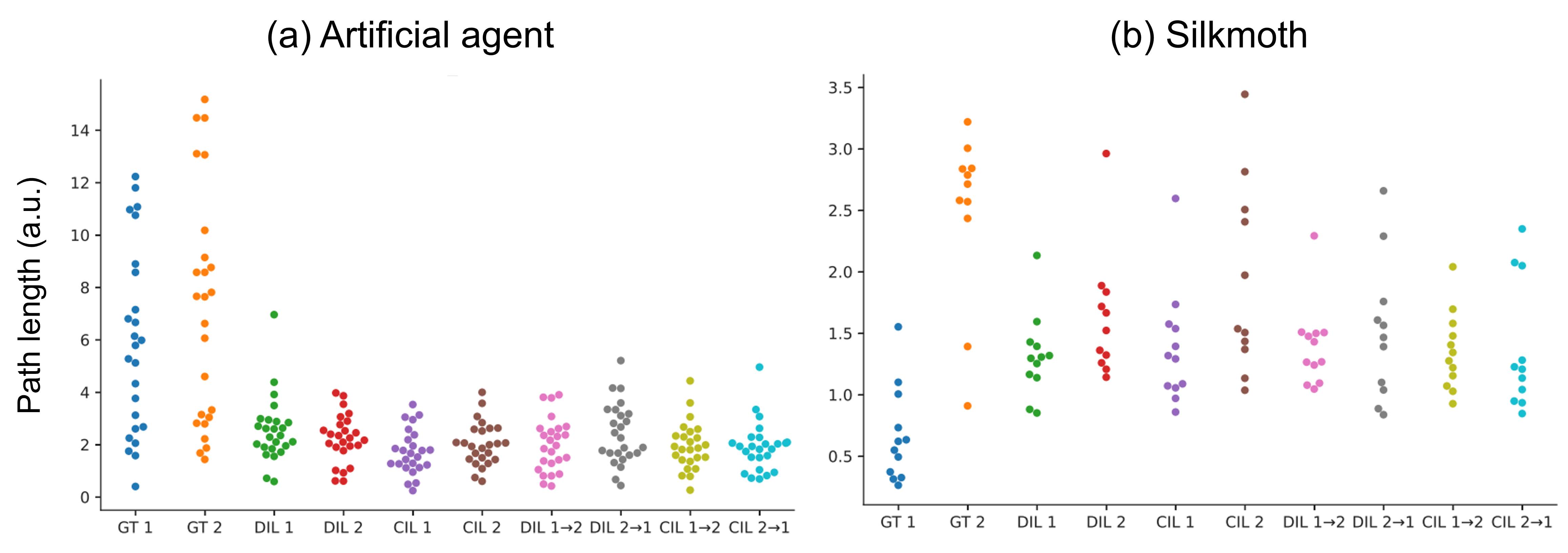}}
\caption{{\bf Counterfactual path length analysis.}
(a) For the artificial agents task, each dot is the mean path length of the chasers in one episode. GT 1 and GT 2 are measured behaviors under the no-sharing (1) and sharing (2) rewards for chasers. DIL 1, DIL 2, CIL 1, and CIL 2 are DQDIL and DQCIL predictions for conditions 1 and 2, respectively.  DIL and CIL ``1 $\rightarrow$ 2'' and ``2 $\rightarrow$ 1'' denote DQCIL and DQCIL counterfactual queries in which the model is fed the opposite condition cue. (b) Silkmoth results with the same layout are shown. Condition 1 is full sensory input and Condition 2 is partial (wind and vision sensors are blocked). 
}
\label{fig:CF_results}
\end{figure} 

\begin{figure}[!t]
\centerline{\includegraphics[width=1 \textwidth]{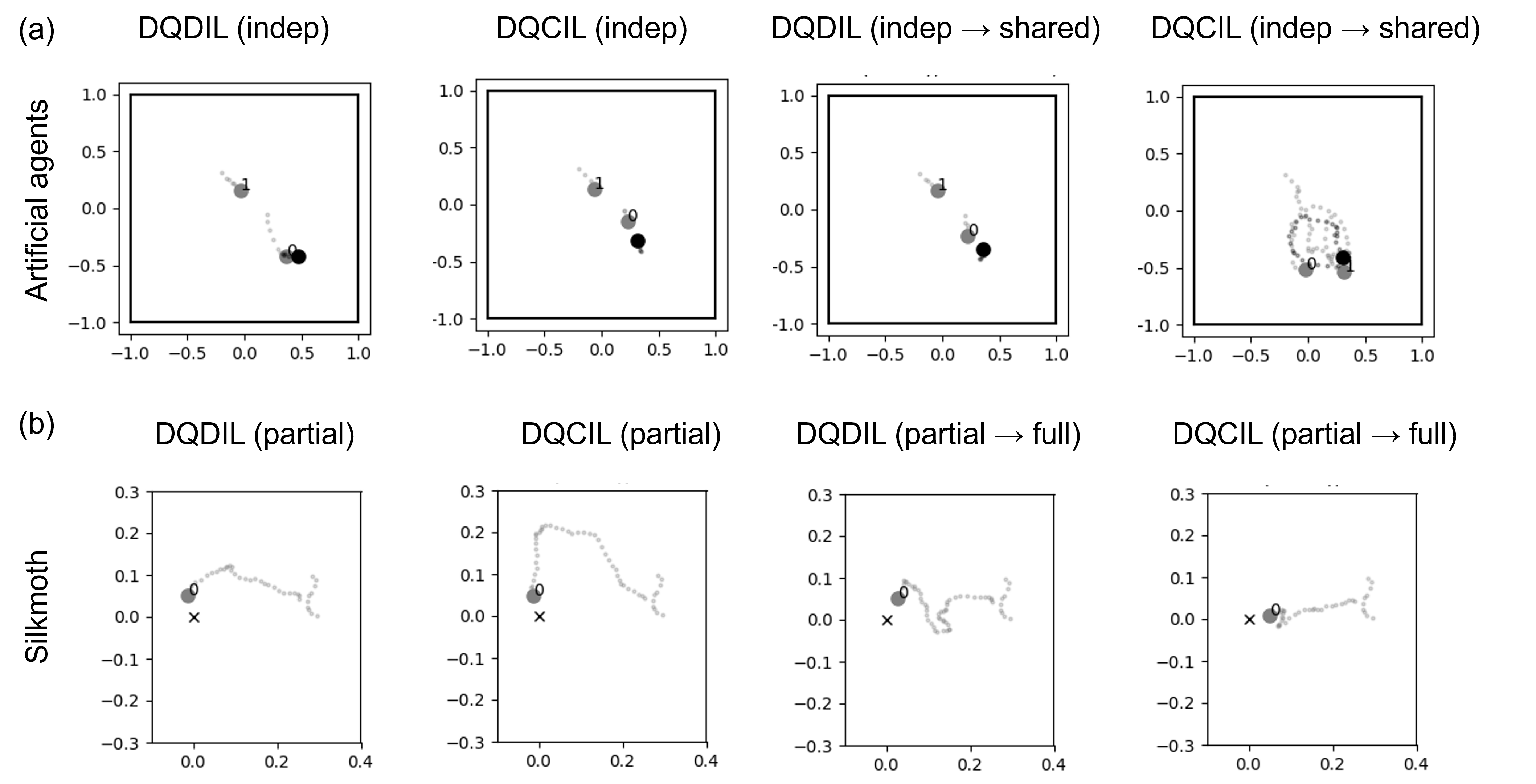}}
\caption{{\bf Counterfactual trajectory examples for (a) artificial agents and (b) silkmoth.}
The layouts for each subfigure are the same as Fig. \ref{fig:example}. For agents, from left to right, each column indicates DQDIL and DQCIL predictions with independent reward (indep: Condition 1); DQDIL and DQCIL counterfactual predictions with shared reward queries (shared: Condition 2) in the Condition 1 setting.
For silkmoth, each column indicates DQDIL and DQCIL predictions with no wind and visual sensor (partial: Condition 2); DQDIL and DQCIL counterfactual predictions with full sensory queries (full: Condition 1) in the Condition 2 setting.
}
\label{fig:CF_example}
\end{figure} 

In the artificial agent counterfactual experiment, as shown in Fig. \ref{fig:CF_results}a, the GT episodes show no reliable distance gap between the two reward conditions: the bootstrap 95\% CI for the Condition 1 minus Condition 2 difference spanned ‐3.407 to 0.959, but DQCIL reproduced those distances more faithfully than DQDIL. Relative to GT, the median absolute errors were 1.003 for CIL (95\% CI [-1.212, 3.265]) and 1.525 for DIL ([-0.678 to 3.742]). 
When we queried the models with the opposite condition cue, only the DQCIL prediction for ``1 $\rightarrow$ 2'' obtained a negative distance shift (95\% CI [-0.927, -0.017]), matching the hypothesis that chasers travel farther when they switch from individual to shared reward, whereas all other counterfactual contrasts had intervals overlapping zero. 
These results indicate that the counterfactual head improves fidelity to both observed and hypothetical conditions, especially for the crucial individual-to-shared transition.
In Fig. \ref{fig:CF_example}a, we show the example of the counterfactual model, which predicts longer chaser paths when the binary cue is switched from individual to shared reward.

For the silkmoth experiment in Fig. \ref{fig:CF_results}b, GT episodes show a clear increase in path length when sensory input was reduced (Condition 2), with the bootstrap 95\% CI for the Condition 1 minus Condition 2 difference entirely negative ([-2.216 to -1.340]). DQCIL reproduced this drop ([-1.055 to -0.054]), whereas DQDIL failed ([-0.667 to 0.011]). Relative to GT, CIL’s median absolute error was 1.274 (95\% CI [0.585, 1.926]), slightly smaller than DIL’s 1.514 ([0.936, 2.026]). In counterfactual queries, only DQCIL’s ``2 $\rightarrow$ 1'' prediction switching from partial to full sensory input show the expected increase in distance (95\%CI [0.018, 1.104]), while all other contrasts had intervals that overlapped zero, highlighting DQCIL’s unique ability to capture the empirically observed sensitivity to sensory deprivation.
In Fig. \ref{fig:CF_example}b, we show the example of the counterfactual model, which predicts a longer silkmoth path when the binary cue is switched from partial to full sensory conditions. 

\section*{Discussion}
The objective of this paper was to understand how data-driven RL locomotion models influence the fidelity and adaptability across both artificial and biological agents.
Unlike classical rule-based formulations that hard-code local interaction rules \cite{reynolds1987flocks,Vicsek95}
and black-box trajectory models trained either by supervised sequence learning \cite{Eyjolfsdottir17,fujii2024decentralized,teshima2025flight} or deep RL \cite{peng2020learning,zhu2022learning}, our framework first recover an explicit locomotion model characterized by a damping coefficient and a discrete thrust amplitude directly from demonstration trajectories and then embedding that model inside a RL loop. 
This locomotion parameter identification step obtained smaller errors for newts and silkmoths but noticeably larger errors for artificial agents and flies (Table \ref{tab:param_estimates}).  
Combined with Fig. \ref{fig:velocity}, we attribute the higher error for agents to the integration of high-speed segments that amplify discretization noise, and the poorest fly fit to its bimodal velocity histogram, which forces any single-mode parametric model to trade off between an excess of near-zero points and a long high-speed tail.  
Crucially, these estimation trends may affect the subsequent RL results (Fig. \ref{fig:models}): species with tighter parameter fits (newts, silkmoths) partially support our hypotheses that our approaches were better than the previous ones in reward acquisition and trajectory similarity, whereas flies remain the most challenging, which is discussed in the next paragraph on policy evaluation.

For policy evaluation, DQAAS \cite{fujii2024adaptive} excelled in the fly domain because the task may sometimes offer short ballistic steps derived from pauses toward a female.
Moreover, with the smallest dataset (107 episodes) in our study, the per-step supervised gradients injected by DQAAS reduced variance and prevented overfitting to the long-tail of rare high-velocity spikes, while the reward signal in DQDILs was largely redundant because even simple policies nearly always contact the target. 
By contrast, on newts and silkmoths, DQDIL partly outperformed DQAAS, where male newts often stop entirely and silkmoths zig-zag while casting through an odor source. 
Our DQDIL would be unaffected by temporal jitter that would penalize a strictly step-aligned loss. 
We also note that both species supply hundreds of trajectories (280 and 240 for newts and silkmoth, respectively), the model can learn a stable DTW-based reward surface, making trajectory-level shaping a stronger learning signal than per-step labels in DQAAS \cite{fujii2024adaptive}. In short, DQAAS can be advantageous when optimal behavior is brief, well aligned, and data are scarce (flies), whereas DQDIL would perform well when we examine matching global spatiotemporal structure rather than individual micro actions (newts and silkmoths).

In counterfactual analysis (Figs. \ref{fig:CF_results} and \ref{fig:CF_example}), our approach with DQCIL proved effective in artificial agent and silkmoth domains, but the effectiveness was limited to specific training conditions and queries. 
For artificial agents, adding the pre-trained prior lets DQCIL generate trajectories with increased path length when asked to flip from no-reward-sharing to reward-sharing, whereas the reverse query did not affect the path length.
Since the agent dataset was sparse and dominated by high-speed maneuvers, we retained a pre-training prior so that the counterfactual head could reflect the reward-sharing dynamics while the reverse query remained under-constrained.
In silkmoths, we omitted the pre-training in the main text, but the model still produced increased path length like GT when queried from the partial sensory condition to the full sensory condition.
We speculate that a large, low-speed dataset already anchored the model, and that only the sensory-partial-to-full flip was well posed, whereas the opposite direction lacked the requisite stimulus information. 
These results highlight that our counterfactual module, although entirely data-driven, can exploit prior knowledge when available and could be made even more accurate by encoding fully known or partially known dynamics directly into the network’s latent state or loss structure.
Since the framework does not rely on a hand-coded mathematical model but still provides interpretable locomotion parameters and state-action values, it would also be well suited  as such as wild animals and a controllable in-silico laboratory for virtual experiments whose behaviors is difficult to formalize.

In conclusion, we introduced AnimaRL, a three-stage framework that first fits an explicit locomotion model from demonstration trajectories and then optimizes an offline and online deep-RL policy, thereby bridging the Real-to-Sim gap for agents with unknown dynamics.  
Across four different domains (synthetic chasers, flies, newts, and silkmoths), the distance-based RL model (DQDIL) matched or exceeded prior baselines in trajectory fidelity, while its counterfactual extension (DQCIL) partly predicted how path length would change under different conditions.  
These results suggest both the promise and the current limitation of the approach. For the former, our results elevate Real-to-Sim fidelity to a level that complements existing Sim-to-Real work \cite{rusu2017sim}, thereby laying a balanced foundation for closed Real-Sim-Real loops that can drive both mechanistic insight and applied experimentation. For future work, we can consider embedded mixture or state-switching dynamics for bimodal movers such as flies, incorporate partially known physics directly into the latent loss, and scale the framework to three-dimensional settings.

\section*{Materials and Methods} 
\subsection*{Datasets.}
Here we describe the demonstration datasets of artificial agents, flies, newts, silkmoth for AnimaRL framework.  
Each of these four datasets shares some common features: first, all datasets consist of coordinate data in a two-dimensional plane, with tracking recorded as agents navigate this space. Additionally, each dataset assumes that touching a designated target (e.g., a prey, a female, a source) provides a reward for the pursuer. Due to varying circumstances surrounding data collection, the sample sizes differ between datasets. However, in each case, we have split the data into training and test sets to verify generalization performance. In particular, when testing for online policy adjustments, we ensured that none of the locomotion parameter estimations or offline policy learning data used in training was included in the test data.

\subsubsection*{Artificial agent dataset.}
We conducted our simulation in a two-dimensional chase-and-escape environment that mirrors the real-world datasets analyzed later in this paper. The environment is a customized version of the ``predator-prey'' scenario from the Multi-Agent Particle Environment (MAPE) \cite{lowe2017multi,tsutsui2022emergence,tsutsui2024collaborative}. Following the configuration in \cite{tsutsui2024collaborative}, the play field spans [-1, 1] on both the x- and y-axes (see Figs. \ref{fig:diagram} and \ref{fig:example}). All agents are discs of diameter 0.1, there are no obstacles, and contacts between predators are ignored for simplicity. Dynamics evolve in discrete time with a step of 0.1 s, and each episode terminates after 30 s. Initial positions are sampled uniformly from the square [-0.5, 0.5]. An episode ends in favor of the predators if either predator touches the prey within the time limit; otherwise, the prey wins. If any agent leaves the arena, its opponent(s) are declared the winners. Each agent can choose from 13 actions-acceleration in 12 equally spaced directions (every 30 degrees in its local frame) or no acceleration. To study domain-adaptation effects, prey mobility is identical in the source and target environments, whereas predator mobility is set to 120\% of the prey’s in each domain. 

We obtained the demonstration data from the original study \cite{tsutsui2024collaborative} with deep RL with individual and shared reward conditions. 
We obtained 500 episodes for demonstration with randomized initial locations (maximum: 17.9 s, minimum: 0.2 s, and total: 2983.8 s).
We split the datasets into 400, 50, and 50 for training, validation, and testing (200, 25, and 25 in each condition) during locomotion parameter estimation and offline policy learning, respectively.
For the online policy adjustment, we used 50 train and 10 test episodes from the above 100 episodes for the validation and testing data.

The two predators captured the prey at rates of 0.424 and 0.248 for the individual reward condition, and 0.814 for the shared reward condition, respectively (there were some cases of capturing at the same time for two predators). These are defined as reward ($+1$) for predators, and the duration [s] is defined as the prey's reward.
The state was defined by the absolute positions and velocities of all individuals, while the observations were defined by each individual's absolute position, as well as the relative positions and velocities of others.

\subsubsection*{Fly dataset.}
In the fly dataset, we used two male flies and one female fly.
Male flies actively pursue females, but do not pursue other males, as known in previous studies \cite{demir2005fruitless}. 
A wild-type strain of \textit{Drosophila melanogaster}, Canton-S, was used for the experiments. The fly strain was raised in standard cornmeal yeast medium at 25 $\pm$ 1 $^\circ$C and 40\%–60\% relative humidity in the 12 h / 12 h light / dark cycle. Adult flies were collected within 6 hours after eclosion to obtain virgin flies. Male flies were housed individually to improve courtship motivation, while female flies were housed in groups of approximately 10–20 individuals. Flies aged 4 to 8 days were used for video recording. Two male flies and one female flies were introduced into an observation chamber with sloped walls (60 mm in diameter, 3.50 mm in depth), and their behavior was recorded for 30 minutes at 30 fps using a CMOS camera (DFK 33UP1300, The Imaging Source Asia Co., Ltd.) equipped with a zoom lens (M0814-MP2, CBC Optics Co., Ltd.). The chamber was illuminated by white LED light during recording.

After obtaining the data, we estimated the range of movement for the flies based on the data and scaled the coordinates within a range of -1 to 1. 
We split the original video file based on the distance thresholds between the male and female flies, with a distance threshold of 0.3 for the episode start and 0.2 for the episode end (assuming a contact between male and female flies). Using these criteria, we identified a total of 107 episodes (maximum: 47.8 s, minimum: 0.4 s, total: 425.6 s).
The two males approached the female within the threshold at rates of 0.720 and 0.280, respectively.
The original video was sampled at a frequency of 30 Hz and down-sampled to 10 Hz. 
We split this limited dataset into 94 and 13 episodes for locomotion parameter estimation and its validation, respectively. 
For offline policy learning, we split the dataset into 84, 13, and 10 episodes for training, validation, and test, respectively.
For the online policy adjustment, we used 50 train and 10 test episodes, where the test episodes are the same as those of offline policy learning.
The definitions of rewards, states, and observations are the same as the artificial agents.

\subsubsection*{Newt dataset.}
Similar to the fly dataset, the experiments involving newts were conducted using a configuration of two males and one female. It is known that female newts release pheromones, and that males respond to these chemical cues \cite{imorin}. In this dataset, each behavioral episode was defined as a sequence in which a male, starting from a certain distance away, approached a female and attempted to make physical contact.
The experiments were conducted using Japanese fire-bellied newts (\textit{Cynops pyrrhogaster}). During mating season, males gather in aquatic environments such as ponds and rice paddies, where they engage in mating behavior with visiting females. The individuals used in this study were captured in the wild in September 2020 and were kept in plastic tanks (26 cm $\times$ 43 cm $\times$  15 cm) for over six months prior to the experiments, which took place from April to June 2020.
In each trial, two males and one female were transferred to an experimental tank, and their behaviors were recorded on video for a duration of 30 minutes. To ensure that both sexes exhibited high reproductive motivation, only the ten trials in which at least one courtship display (i.e., tail-waving dance) by a male was observed during the 30-minute session were selected for analysis.
The positions of males and a female in the video recordings were determined using UMAtracker \cite{umatracker} combined with manual correction. The inter-individual distance between males and the female was calculated, and a behavioral episode was defined as a sequence in which a male approached a female from beyond a threshold distance to close proximity. Episodes in which this sequence exceeded 300 seconds or did not result in close proximity were excluded from analysis.

After obtaining the data, we scaled the coordinates within a range of -1 to 1. 
One of the authors manually annotated the onset of courtship behavior by visually watching the videos, and extracted each courtship event. If the interval between courtship events was too long, we extracted a 50-second episode ending at the courtship onset.
We split the original video file based on the distance thresholds between the male and female newts, with a distance threshold of 1.0 for the episode start and 0.15 for the episode end (assuming a contact between male and female newts). 
Using these criteria, we identified a total of 280 episodes (maximum: 50 s, minimum: 2.5 s, and total: 12141.6 s).
The two males approached the female and engaged in courtship behavior at rates of 0.507 and 0.442, respectively.
The original video was sampled at a frequency of 30 Hz and down-sampled to 10 Hz. 
We split this limited dataset into 240 and 40 episodes for locomotion parameter estimation and its validation, respectively. 
For offline policy learning, we split the dataset into 240, 20, and 20 episodes for training, validation, and test, respectively.
For the online policy adjustment, we used 50 train and 10 test episodes, where the test episodes are included in those of offline policy learning and the validation episodes in locomotion parameter estimation.
The definitions of rewards, states, and observations are the same as the artificial agents.

\subsubsection*{Silkmoth dataset.}
Different from the above three datasets, we used a silkmoth moving toward an odor source for silkmoth dataset. 
A male silkmoth (Bombyx mori) behavior was collected using a custom-designed virtual reality (VR) system \cite{yamada2021multisensory} tailored to simulate a naturalistic environment. The VR system provided a multi-sensory stimulus setup, including a pheromone-based odor stimulator, an optical flow-based visual stimulator, and a wind generator to mimic environmental cues. The odor field was simulated using smoke emitted in a darkroom, with its spread visualized through a laser sheet and recorded by a high-sensitivity camera, creating a virtual odor plume based on real diffusion patterns. This setup enabled the moths to respond to directional cues in a manner consistent with free-walking behaviors observed in real-world environments. Additionally, the system included an LED array to create optical flow, mimicking natural visual cues, and a push-pull rectifier for generating wind. Each stimulator's effectiveness was verified through behavioral responses, ensuring that the VR system accurately replicated real-world conditions for tracking and analyzing silkmoth search behavior. 

The measurement setup was the same as the previous study \cite{yamada2021multisensory}, but one of the authors newly measured the behavior in the condition i and 1 (i.e., with and without wind and vision input, respectively) 6 episodes with 5 odor patterns (in total, 60 episodes). 
We redefine the condition names as full-sensory and reduced-sensory conditions. 
Similarly to the previous study \cite{yamada2021multisensory}, episodes exceeding 300 s were considered failures.
Minimum length of episodes was 9.0 s, and total: 6611.5 s.
The behavioral dataset was sampled at a frequency of 30 Hz and down-sampled to 2 Hz. 
We split this limited dataset into 36 and 24 episodes for locomotion parameter estimation and its validation, respectively. 
For offline policy learning, we split the dataset into 36, 12, and 12 episodes for training, validation, and test, respectively.
For the online policy adjustment, we used 36 train and 24 test episodes, where the test episodes are included in those of offline policy learning and the validation episodes in locomotion parameter estimation.
For each data split, we used 3 odor patterns for training and the remainig 2 patterns for validation or testing.  

A silkmoth in full-sensory and reduced-sensory conditions reached the odor source at rates of 1.000 and 0.633, respectively.
These are defined as reward ($+1$) for a silkmoth.
The state was defined as body angle, odor from left and right (or both) antennae, wind from four directions (front, back, left, or right), vision (left or right) in addition to the absolute position and velocity.
The observations were defined as the state except for the absolute position because the task will be too easy to accomplish if the ego-position is known.

\subsection*{Background of reinforcement learning.}
We study a sequential, fully observable multi-agent decision process, formalized as the tuple $(K,S,A,\mathcal{T},R,\gamma)$ illustrated in Fig. \ref{fig:diagram}. Here, $K$ denotes the fixed number of agents; $S$ is the set of global states $s$; $A=[A_1,\ldots,A_K]$ is the joint-action space with joint action $\avec\in A$ and local action $a_k\in A_k$ for agent $k$; $\mathcal{T}(s'\mid s,\avec):S\times A\times S\to[0,1]$ gives the transition dynamics; $R=[R_1,\ldots,R_K]:S\times A\to\mathbb R^{K}$ is the joint reward; and $\gamma\in(0,1]$ is the discount factor. Each agent $k$ seeks a policy $\pi_k:S_k\times A_k\to[0,1]$ that maximizes its expected discounted return $G_k=\sum_{t=0}^{T}\gamma^{t} R_{k,t}$ over the horizon $T$. The action-value of following $\pi_k$ from state–action pair $(s_k,a_k)$ is $Q^{\pi_k}_k(s_k,a_k)$, and the optimal joint action–value function $Q^{*}(s,a)$, which attains the maximum over all policies in every state, satisfies the Bellman optimality condition:

\begin{equation}
Q_k^*(s_k,a_k) = \E\left[{R_k(s_k,a_k) + \gamma \sum_{s_k'} \mathcal{T}_k(s_k'|s_k,a_k) \max_{a_k'} Q_k^*(s_k',a_k')}\right].
\label{eq:bellman}
\end{equation}

\noindent
Here $\mathcal{T}_{k}$ denotes the transition model of agent $k$, and the corresponding greedy policy is $\pi_{k}(s_{k})=\argmax_{a_{k}\in A} Q^{*}_{k}(s_{k},a_{k})$.
Since biological agents tend to behave independently rather than in central control, we consider each agent to maintain an independent policy network.

In realistic multi-agent domains, such as collective animal motion, the true transition law
$\mathcal{T}^{E}(s'^{E} \mid s^{E},\avec^{E})$
(i.e., the underlying locomotion parameters in this study) is rarely known in closed form.
Instead, one can infer it from demonstration data, for example, recorded trajectories of the group.  
In this work, we carry out online RL under an assumed model $\mathcal{T}(s' \mid s,\avec)$.
We are fully aware that it differs from $\mathcal{T}^{E}$, but we estimate the locomotion parameters from the demonstration data.

\subsubsection*{Deep Q-Network (DQN).} 
For clarity, we now revert to a single-agent notation. DQN\cite{mnih2015human} approximates the action-value function $Q(s,a)$ by a deep network $Q(s,\cdot;\theta)$ with parameters $\theta$.  
A separate target network is synchronized with the online network every $\tau$ steps to stabilize the temporal-difference targets.  Experiences are accumulated in a replay buffer $\mathcal{D}^{\text{replay}}$ and drawn uniformly for mini-batch updates.

We actually used a double Q-learning framework \cite{van2016deep} improving standard Q-learning by addressing the issue of overestimation in action values. This is achieved by using two networks: one network computes the best next action (by taking the argmax over the subsequent state values), while the other network, known as the target network, is used to estimate the value of this chosen action, reducing the upward bias typically found in traditional Q-learning updates \cite{van2016deep}. 
The double Q learning temporal difference loss is as follows:
\begin{equation}
    J_{DQ}(Q) = \sum^{T-1}_t \left( R_t + \gamma Q(s_{t+1}, a^{\max}_{t+1}; \theta') - Q(s_t, a_t; \theta) \right)^2,   
\label{eq:ddqn}
\end{equation}
where $\theta'$ refers to the parameters of the target network, and $a^{\max}_{t+1} = \argmax_{a_{t+1}} Q(s_{t+1}, a_{t+1}; \theta)$. 
Additionally, prioritized experience replay \cite{schaul2016prioritized} has been introduced to enhance learning efficiency by sampling more critical transitions more frequently from the replay buffer, allowing the model to focus on learning from significant experiences.

\subsection*{Reinforcement learning environment.}
Here we describe the requirements for simulating agent behavior.
The simulation setup requires three main components: the agent velocity transition function (as described in Eq. \ref{eq:velocity_transition}, with parameter estimation covered in the following section), the termination conditions, and the reward settings. 

Only for the silkmoth simulations, it is important to account not only for the transition model of the agent’s velocity but also for the transitions of the odor, visual, and wind patterns. To facilitate this, we developed a simplified version of the simulation framework, incorporating pattern images for odor and wind generated from experimental measurements (previously developed by one of the authors \cite{yamada2021multisensory}). This simplified simulator enables output of the next time step's odor, visual, and wind values. Additionally, the silkmoth’s body orientation was approximated by the direction of its velocity vector for computational simplicity.

The termination conditions are based on whether the agent reaches the target (prey in artificial agents, female individuals in the fly and newt datasets, or an odor source in the silkmoth dataset), exceeds the designated time limit based on the maximal time length of observations (artificial agents: 14.8 s, flies: 50 s, newts: 50 s, silkmoths: 300 s), or crosses the virtual boundary. For artificial agents and newts, this boundary is defined as a square with an absolute value of 1.1 for both the x and y coordinates (forming a virtual boundary within a $2 \times 2$ square). For flies, the boundary is defined as a circle with a radius of 1.1 (enclosing a virtual circular boundary with radius 1), while for silkmoths, no explicit boundary is set.

For the rewards, the pursuing agent (predator or male) in artificial agents, flies, and newts received a reward of $+1$ when they reached the target (prey or female). Conversely, the target received a reward proportional to the time elapsed, as they are typically caught within a limited time. Additionally, a penalty of $-10$ was imposed if any agent crossed the predefined boundary. For the silkmoth, a reward of $+1$ was given upon reaching the odor source.

\subsection*{Locomotion parameter estimation.}
Here we describe the approach used for locomotion parameter estimation, but the core formulation was introduced in the Results section. 
From each absolute velocity, we first detect the set of rest-to-motion transitions,
\begin{equation}
\mathcal I = \bigl\{t \mid |v|_{t}<\varepsilon, |v|_{t+1}\ge\varepsilon\bigr\},
\qquad 
\varepsilon = th_{acc}\Delta t, 
\end{equation}
where $th_{acc}$ is a threshold of acceleration.
Let $|v|_{\text{on}} = \mathrm{median}\{|v|_{t+1}\mid t\in\mathcal I\}$ be the typical velocity immediately after such a transition and  
$|v|_{\max} = \mathrm{median}\{|v|_t\mid v_t>P_{99}(|v|)\}$ the median of the upper \(1\%\) of all velocities.  
We then set
\begin{equation}
u = \frac{|v|_{\text{on}}}{\Delta t},
\qquad
d = \frac{|v|_{\text{on}}}{|v|_{\max}},
\end{equation}
so that $u$ approximates the characteristic acceleration ($|v|_{\text{on}}$ per sampling interval $1/\Delta t$), while $d$ captures the ratio between that onset speed and the sustained top-speed, serving as an empirical damping coefficient.

Next, we shortly summarize the validation methodology and the corresponding outcomes. Verification of locomotion parameter estimation was performed by evaluating the root mean square error (RMSE) between the predicted and actual velocities.
Estimated damping parameters and amplitudes of discrete inputs are shown in Fig. \ref{fig:example} (The ground truth of them is only in the artificial agents).
The RMSE values were shown in Table \ref{tab:param_estimates}.

\subsection*{Offline and online policy learning (DQDIL and DQCIL)}
Here we explain how we perform pre-training and online fine-tuning to train DQDIL on demonstrations and then augment it with the counterfactual head (DQCIL). 
In numerous practical reinforcement-learning settings, observation data produced by multi-agent systems are available, whereas reliable descriptions of the underlying dynamics are not.  Accordingly, an agent must leverage such demonstration data to narrow the behavioral gap between the source domain and the target environment.
According to the deep Q-learning from demonstrations (DQfD) approach \cite{hester2018deep}, we use two primary steps: offline policy learning (pre-training), where the agent learns to imitate the demonstrator, and subsequent training in the RL environment using actions sampled from the pre-trained model. The network is updated with the 1-step double Q-learning loss in Eq. (\ref{eq:ddqn}), which aligns the model with the Bellman equation and serves as a foundation for TD learning. Additionally, action-supervised loss \cite{nakahara2023action,fujii2024adaptive} or imitation rewards \cite{fickinger2022cross} are employed to enhance pre-training efficiency.

In the previous DQAAS \cite{fujii2024adaptive}, in a similar setting of chase-and-escape and football tasks, the action supervision loss worked well compared with the large margin classification loss \cite{hester2018deep}. 
The supervised loss is crucial for pre-training because the demonstration data usually covers a narrow part of the state space and does not take all possible actions \cite{fujii2024adaptive,nakahara2023action}.
Here we consider a single agent case for simplicity (i.e., we removed the notation of agent $k$, but we can easily extend it to multi-agent cases).

\subsubsection*{Distance-based pseudo reward.}
Distance-based pseudo-rewards have been explored through various methodologies in imitation learning and reinforcement learning, each tailored to specific challenges in aligning agent behaviors with expert demonstrations. 
Optimal transport metrics, such as Wasserstein \cite{dadashi2021primal},  Sinkhorn \cite{papagiannis2022imitation}, and Gromov-Wasserstein distances \cite{fickinger2022cross}, have been widely employed to match occupancy measures across domains or metric spaces, enabling sample-efficient learning while avoiding adversarial objectives.
Since our objective is to preserve the fine-grained temporal ordering of each trajectory rather than merely matching marginal occupancy distributions, we prefer DTW \cite{vintsyuk1968speech,sakoe1978dynamic}, which aligns sequences frame-by-frame and thus captures trajectory variations.
Previous work used DTW for the alignment of trajectories through a latent space embedding, facilitating robust imitation learning under domain discrepancies \cite{gupta2022learning}. However, our method feeds the raw DTW distance directly into the reward, ensuring that the policy optimizes the actual temporal correspondence of the observed motions.
Our approach avoids extra embedding hyperparameters, simple and interpretable implementation, and would be better suited to reproducing real-world biological movements with a smaller amount of demonstrations.

Therefore, in this study, we reshape the environment reward by linearly mixing the original target-touch signal $R^{\text{touch}}_t$ with a DTW-based trajectory similarity
\begin{equation}
R_t = R^{\text{touch}}_t -\alpha R^{\text{DTW}}_t,
\label{eq:r_mix}
\end{equation}
where $\alpha$ is a weighting factor.  
Here, we consider the simulated and corresponding demonstrated (expert) state histories be $s = s_1,\ldots ,s_t,\ldots ,s_n$ and $s^E = s^E_1,\ldots ,s^E_j,\ldots ,s^E_m$, with lengths $n$ and $m$, respectively. DTW \cite{vintsyuk1968speech,sakoe1978dynamic} aligns these two sequences by searching for a \emph{warping path} $W(s,s^E) \in \mathbb R^{n\times m}$ that accumulates the smallest possible cost subject to the usual monotonicity, continuity, and boundary constraints: the $(t,j)$ element $W(s,s^E)_{t,j}$ records the minimum summed local distance $d(s_t,s^E_j)$ (here Euclidean) over all admissible paths that pass through the grid position $(t,j)$. Accordingly, we set $R^{\text{DTW}}_t = \min_{1 \le j \le m} W(s,s^{E})_{t,j}$ which defines the DTW penalty at time step \(t\) as the smallest cumulative alignment cost between the current simulated state \(s_t\) and every expert state \(s^{E}_j\).

For distance-guided RL we use the standard one-step double-Q loss and a small $\ell_2$ penalty.
The total loss used to update the network is as follows:
\begin{equation}
  J(Q) = J_{DQ}(Q) + {\lambda_1}J_{\ell_2}(Q).
  \label{eq:losses}
\end{equation}
The shaped reward in Eq. (\ref{eq:r_mix}) already embeds the DTW imitation signal in the return.
The same shaped reward $\tilde R_t$ is used during online fine-tuning in DQDIL. 

\subsubsection*{Counterfactual imitation learning (DQCIL).}
DQDIL imitates trajectories that were actually observed, but it cannot generalize to ``what-if'' scenarios in which those conditions change. In other words, the latent state that DQDIL learns is entangled with the treatment variable (reward-sharing on/off, sensory input full/partial in agents and silkmoth, respectively), and when the condition flag is flipped at test time, the network simply receives an out-of-distribution input and produces unreliable roll-outs.
To overcome this problem, we borrow the idea of adversarial treatment-prediction from counterfactual inference and domain-invariant representation learning \cite{ganin2016domain, johansson2016learning, shalit2017estimating}: a small auxiliary head tries to recover the binary condition from the shared features while a gradient-reversal layer forces those features to hide that information. 
By coupling this loss with the original DTW-shaped objective, we obtain DQCIL, which keeps the policy optimal for the observed condition but embeds enough dynamics to synthesize plausible trajectories when the condition is hypothetically switched.

We assume that each transition carries a binary treatment flag $c_t \in {0,1}$ indicating the experimental condition (e.g., reward sharing on/off or full/partial sensory input). To encourage the Q-network to produce condition-invariant state features while still exposing the flag for counterfactual queries, we attach a treatment-prediction head $\psi$ that outputs $\hat c_t=\psi(h_t)$ from the shared hidden representation $h_t$ and train it with the cross-entropy loss:
\begin{equation}
J_{\text{tr}} = -\Bigl(
      c_t\log\hat c_t + (1-c_t)\log\bigl(1-\hat c_t\bigr)
 \Bigr).
\end{equation}

During back-propagation, the gradient flowing from $J_{\text{tr}}$ through $h_t$ is multiplied by $-1$ (gradient-reversal layer \cite{ganin2016domain}), so the encoder learns to remove causal information about $c_t$ while the prediction head learns to recover it. 
The full objective therefore becomes
\begin{equation}
J_{\mathrm{DQCIL}}(Q) = 
J_{\text{DQ}}(Q) + \lambda_1\,J_{\ell_2}(Q) + \lambda_2\,J_{\text{tr}},
\label{eq:dqcil}
\end{equation}
where $\lambda_2$ controls the strength of the counterfactual regularizer. This design follows the treatment-prediction rationale of the previous work\cite{fujii2022estimating, fujii2024estimating}: by balancing the hidden state with respect to $c_t$, the network can generate trajectories for the observed condition and plausible alternatives when the flag is flipped at inference time.

\subsubsection*{Neural network architectures.}
Both DQDIL and its counterfactual variant DQCIL use the same recurrent neural network (RNN) backbone.
For every agent, raw observations first pass through two fully connected (FC) layers that serve as a shared encoder. The resulting feature vector is fed to a single-layer gated recurrent unit (GRU\cite{Cho14}) whose hidden state $h_t$ is carried across time. 
The decoder is dueling network\cite{wang2016dueling}: a state branch outputs a baseline value $V(h_t)$ while an advantage branch produces $A(h_t,a_t)$, and the two are combined into $Q(h_t,a_t)=V(h_t)+A(h_t,a_t)-\frac 1{|A|}\sum_{a_t'}A(h_t,a_t)$ with two FC layers.
All hidden unit sizes are 32. 
DQCIL augments each agent with a treatment-prediction head (two FC layers, 8 hidden units) that is preceded by a gradient-reversal layer. All heads receive the same GRU output, so parameters upstream of the reversal gate are updated jointly by the TD loss (and the DTW-shaped reward in DQDIL) and the adversarial signal, giving a compact but expressive architecture for both factual and what-if trajectory generation.

\subsection*{Baseline methods}
We compared our approaches with three baselines below. For fair comparison, the network architectures and training details including hyperparameters are the same as DQDIL unless specified below.  

\subsubsection*{Deep Q-Network (DQN).}
The DQN baseline keeps the identical network but drops all supervision on expert actions, relying exclusively on the one-step double-Q loss and $\ell_2$ regularization norm in Eq. (\ref{eq:dqcil}). The replay buffer is populated online from scratch, ensuring that differences reflect the lack of demonstration guidance.

\subsubsection*{Behavioral cloning (BC).}
BC serves as a purely supervised baseline.
No TD term in the loss function or environment interaction was used.
The loss function is action supervision cross-entropy loss and $\ell_2$ regularization norm in Eq. (\ref{eq:dqcil}).

\subsubsection*{Deep Q-learning with adaptive action supervision (DQAAS).}
DQAAS\cite{fujii2024adaptive} augments the TD objective with the adaptive action-supervision loss $J_{\text{AS+DA}}$, yielding the composite objective $J_{\mathrm{DQAAS}} = J_{\text{DQ}} + \lambda_3 J_{\text{AS+DA}} + \lambda_1 J_{\ell_2}$. In practice, this means that, during both pre-training and online fine-tuning, gradients flow not only from Bellman errors but also from cross-entropy terms that align each agent’s action distribution to its dynamically warped expert counterpart, while every other training detail is kept identical to the other methods.

\subsubsection*{Difference between DQAAS and DQDIL.}
Pseudo-rewards and action supervision represent two distinct approaches with unique strengths and limitations for each other. Pseudo-rewards enhance learning by embedding task-specific signals, often derived from environmental metrics, that encourage exploration and goal-directed behavior. This approach is well-suited for scenarios with sparse rewards or undefined action labels, allowing the agent to discover effective strategies through exploration. However, pseudo-rewards require careful design to ensure alignment with the overarching task objectives, as poorly calibrated metrics may lead to suboptimal policies.

Conversely, action supervision directly integrates expert demonstration data, guiding the agent to replicate these actions via supervised losses. This method excels in environments where high-fidelity reproduction of demonstrated behaviors is crucial, as it reduces exploration overhead. However, action supervision may limit the agent’s ability to generalize or adapt to novel scenarios, particularly if the demonstrations are incomplete or suboptimal.

In this study, we hypothesize that the pseudo reward approach may outperform action supervision for tasks requiring extensive exploration or in cases where demonstrations are limited in coverage or quality. To validate this, we systematically evaluate both methods independently, focusing on their performance under varying task complexities and conditions. This investigation aims to identify scenarios where pseudo-rewards are advantageous and to elucidate the boundaries of their applicability compared to action supervision. By isolating the effects of each approach, we aim to provide clearer insights into their respective strengths and limitations.

\subsubsection*{Training details including hyperparameters for all methods.}
For both offline and online training, hyperparameters $\lambda_1$ for $\ell_2$ regularization loss was set to $10^{-5}$. 
In offline RL training, all training runs were limited to 30 epochs and optimized with the default Adam settings, but according to the previous work \cite{fujii2024adaptive} for small demonstration datasets, we used a lower learning rate than the default parameters.
About the artificial agents, $10^{-6}$ was used for the fast, low-magnitude gradients, 
$10^{-5}$ for silkmoths, whose dense but tiny rewards make the TD target slow-converging;  
and $10^{-4}$ for flies and newts, whose slower kinematics tolerate more aggressive updates.
For agents and silkmoth, $\alpha$ for DQDIL and DQCIL and $\lambda_2$ for DQCIL were set to 10, because they give the DTW bonus and the treatment-prediction loss the same order of magnitude as the sparse target-touch reward.
$\lambda_3$ for DQAAS was set to 50 for agents and silkmoth based on the previous work \cite{fujii2024adaptive}.
For flies and newts, we set $\alpha = 0.5$ and $\lambda_3 = 10$, which were smaller than those in agents and silkmoth because flies and newts had higher DTW distance (longer time and more complex trajectory). 
For AAS and BC in all agents and animals, all learning rates were set to $10^{-3}$ (default values). 

In online RL training, all experiments shared a training horizon of $1{,}005{,}000$ environment steps.  
The optimizers were default Adam settings but the learning rates and the $\epsilon$-greedy exploration schedule were tailored like above.
Regarding the learning rate, training without offline RL (pretraining), the same values were used as the above ones of offline training. 
When online training after pretraining, we reduced the learning rates such that $10^{-6}$ for agents, $10^{-5}$ for flies and newts, and $10^{-7}$ for silkmoth.
In the exploration parameter $\epsilon$, for runs that start without pre-training we decay the exploration rate linearly from $\epsilon_{\text{start}}=0.5$ to $\epsilon_{\text{finish}} = 0.3$ for the three biological domains (for the high-speed agent simulation, we set $\epsilon_{\text{start}} = 0.1$ and $\epsilon_{\text{finish}} = 0.1$) after the first $5{\times}10^{4}$ steps and keep it constant thereafter. 
In the test phase, we set $\epsilon_{\text{test}}= 0.5$ for the three biological domains and $\epsilon_{\text{test}}= 0.1$ for the high-speed agent domain.
When the offline model was fine-tuned, we narrowed the exploration window  $\epsilon_{\text{start}} = 0.3$ (agent) or $0.5$ (silkmoth, fly, newt) down to $\epsilon_{\text{finish}} = 0.1$ or $0.3$  so as to exploit the better initial policy. 
For agents and silkmoth, we set $\alpha = 1$, $\lambda_2 = 1$ for DQCIL, and $\lambda_3 = 10$ because once a reasonable policy is loaded from pre-training, the shaped reward and the regularizer only need to guide rather than dominate learning.
For flies we set $\alpha = 0.5$ and $\lambda_3 = 5$, and for newts, we set $\alpha = 1$ and $\lambda_3 = 0.1$ because we consider that the newt movement imitation may be slightly more challenging than flies in addition to the reasons above.

\subsection*{Statistical analysis.}
We performed statistical analysis for model-comparison tests in Fig. \ref{fig:models} and counterfactual tests in Fig. \ref{fig:CF_results}.
For the former, we quantified how well a learned policy (in particular among DQAAS, DQDIL-PT, and DQDIL) reproduces the GT distributions of path length and episode duration and how much it improves on BC in DTW distance for each species. 
Since one of our goal was to judge how closely a learned policy reproduces GT we compared the KDEs of the path length and duration against the corresponding GT KDEs and used the resulting kernel-density distances as distribution similarity measures.
Kernel-density distances were computed from the empirical distributions using the $gaussian_kde$ implementation in SciPy\cite{silverman1986density, virtanen2020scipy}.
Finally, we assessed spatial fidelity with DTW distance between simulated and GT trajectories, testing improvements relative to BC by paired bootstrap resampling. 

Since the episodes are paired with the same GT or BC episodes, we drew $10,000$ paired bootstrap replicates of the metric difference, computed the resulting 95\% CI, and used its sign to judge practical equivalence or advantage. 
The same resamples fed a bootstrap one-way ANOVA on the absolute GT gaps; a strictly positive F-statistic triggered follow-up contrasts, but so as not to inflate family-wise error, we restricted these post-hoc tests to a priori comparisons (e.g., DQDIL vs DQAAS) instead of exhausting all pairings. 
This non-parametric strategy is appropriate because it makes no Gaussian or equal-variance assumptions, preserves the dependence structure of paired observations, and reports both effect size and uncertainty rather than dichotomous $p$-values. 

For the latter, to evaluate whether the counterfactual model (DQCIL) can predict trajectory changes under unobserved experimental condition, we bootstrap-resampled the mean path length per episode under each condition and computed (i) the absolute error relative to GT in the same condidtion and (ii) the signed shift produced when the binary cue was flipped (e.g., from Condition 1 to Condition 2). 
For every resample we estimated these statistics, obtaining a 95\% CI for the GT error and for the counterfactual shift. 
A CI excluding zero indicates that the model either matches the empirical condition difference (realism) or produces a credible counterfactual adjustment (plausibility). 
Using identical resampling for DQDIL provides a fair ablation, and limiting the analysis to the two scientifically motivated flips (from independent to shared reward in agents, and from partial to full sensory input in silkmoths) avoids the multiple-testing inflation that would accompany a full matrix of condition pairs. 
This design follows domain-invariant representation work, treating the condition flag as a treatment whose causal effect is probed through bootstrap inference. 

\bibliography{main}

\section*{Acknowledgments}
This work was supported by JSPS KAKENHI (Grant Numbers 21H04892 and 21H05300) and JST PRESTO (Grant Number JPMJPR20CA).

\section*{Code avaiability}
The code is avaiable at \url{https://github.com/keisuke198619/animarl}.

\section*{Author contributions statement}
K.F. conceived the study.
M.I., R.T., S.S. conducted biological experiments to obtain the data.
K.F., K.T., Y.T., M.I. conducted numerical experiments to obtain the data.
K.F. designed the framework with N.T. and Y.K. guidance. 
K.F. analyzed the results. 
K.F., Y.T., M.I., N.N., R.T., S.S. discussed from biological perspectives. 
All authors wrote the manuscript. 

\section*{Additional information}
The authors declare that they have no competing interests.

\newpage
\section*{}
\vspace{20mm}
\Large{\bf{Supplementary materials. \\
\\

\noindent } }

\vspace{10mm}

\section*{Text 1. Counterfactual prediction results.}
For artificial agents trained without pretraining, path length distributions under Condition 1 (independent reward) and Condition 2 (shared reward) had no significant difference, and counterfactual queries (1 $\rightarrow$ 2 and 2 $\rightarrow$ 1) did not produce significant shifts in path length (bootstrap 95\% CIs overlapped zero; Supplementary Fig. 1a). 
For silkmoths trained with pretraining, path length distributions under full versus reduced sensory input similarly had no significant difference, and counterfactual flips obtained no significant shifts (bootstrap 95\% CIs overlapped zero; Supplementary Fig. 1b).

\begin{figure}[h]
\centerline{\includegraphics[width=1 \textwidth]{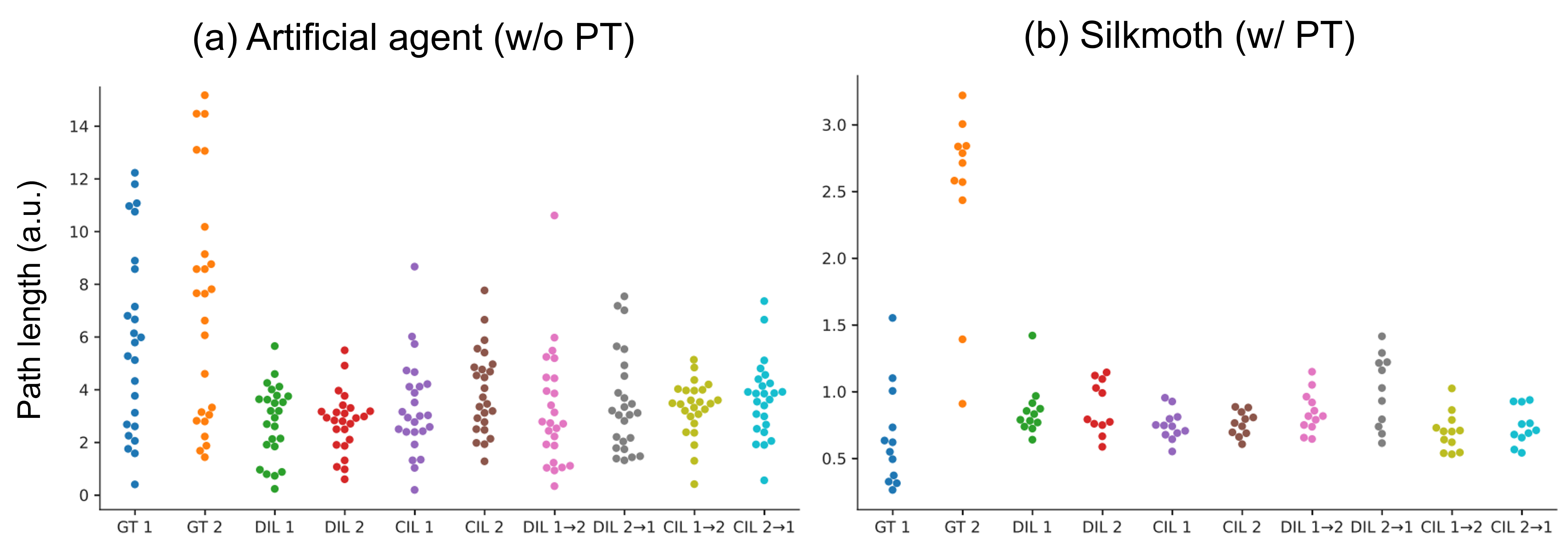}}
\caption{{\bf Counterfactual trajectory examples for (a) artificial agents and (b) silkmoth.}
The layouts for each subfigure are the same as Fig. 5. 
}
\label{fig:CF_example_app}
\end{figure} 
\end{document}